\documentclass{article}
\usepackage{graphicx} 
\usepackage[final]{cpal_2025}
\usepackage{amsmath}
\usepackage{amssymb}
\usepackage{amsfonts}
\usepackage{xcolor}
\usepackage{colortbl}
 \usepackage{booktabs}
\usepackage{enumitem}
\usepackage{amsfonts}
\usepackage{multirow}
\usepackage{hyperref}

\usepackage{url}
\usepackage{subcaption}
\title{ShapLoRA: Allocation of Low-rank Adaption on Large Language Models via Shapley Value Inspired Importance Estimation}

\author{%
  Yi Zhao\textsuperscript{1}\thanks{Corresponding author}, Qinghua Yao\textsuperscript{2},Xinyuan song\textsuperscript{3},Wei Zhu\textsuperscript{4} \\
  \textsuperscript{1}Singapore Management University, \textsuperscript{2}University of Pennsylvania,\textsuperscript{3}Emory University,\textsuperscript{4}University of Hong Kong\\
  \texttt{yizhao@smu.edu.sg, yqinghua@seas.upenn.edu,xinyuan.song@emory.edu,wzhu91@connect.hku.hk}
}

\begin{document}

\maketitle

\begin{abstract}
Low-rank adaption (LoRA) is a representative method in the field of parameter-efficient fine-tuning (PEFT), and is key to Democratizating the modern large language models (LLMs). The vanilla LoRA is implemented with uniform ranks, and the recent literature have found that properly allocating ranks on the LLM backbones results in performance boosts. However, the previous rank allocation methods have limitations since they rely on inexplanable and unreliable importance measures for the LoRA ranks. To address the above issues, we propose the ShapLoRA framework. Inspired by the explanable attribution measure Shapley Value, we combine the sensitivity-based measures with the idea of coalitions in the collaborative games among LoRA ranks, and propose a more explainable importance measure called Shapley sensitivity. In addition, we optimize the workflow of the existing works by: (a) calculating Shapley sensitivity on a separate validation set; (b) Setting up the allocating-retraining procedures for fair comparisons. We have conducted experiments on various challenging tasks, and the experimental results demonstrate that our ShapLoRA method can outperform the recent baselines with comparable tunable parameters.\footnote{Codes and fine-tuned models will be open-sourced to facilitate future research.}
\end{abstract}

\section{Introduction}

Large language models (LLMs) have demonstrated remarkable capabilities, achieving state-of-the-art (SOTA) performance across diverse natural language processing (NLP) tasks \cite{qin2023chatgpt, PromptCBLUE} as well as specialized evaluation benchmarks \cite{huang2023c, li2023cmmlu}, including domain-specific question answering, mathematical reasoning, safety alignment, and instruction comprehension. While LLMs increasingly serve as general-purpose problem solvers, fine-tuning remains critical for optimizing inference efficiency and tailoring the style or tone of model outputs. Recent developments, such as OpenAI’s fine-tuning APIs for GPT-3.5-turbo and GPT-4\footnote{\url{https://openai.com/blog/gpt-3-5-turbo-fine-tuning-and-api-updates}}, underscore its practical relevance. Nevertheless, full-parameter fine-tuning of LLMs is often prohibitively expensive, demanding substantial GPU memory and computational resources not only during training but also during deployment. To address this, parameter-efficient fine-tuning (PEFT) methods \cite{Zhang2023LearnedAA, 2023arXiv230318223Z} have gained prominence, enabling effective adaptation with tunable parameters typically accounting for less than 1\% of the total model weights while drastically reducing training costs.

Parameter-Efficient Fine-Tuning (PEFT) methods have become indispensable for adapting large language models (LLMs). Among these, Low-Rank Adaptation (LoRA) \cite{hu2021lora} has emerged as a particularly effective reparameterization-based approach, achieving widespread adoption in LLM fine-tuning \cite{Xu2023ParameterEfficientFM,Ding2022DeltaTA,Xin2024ParameterEfficientFF}. However, the vanilla LoRA applies a uniform LoRA rank allocation setting across the Transformer layers and linear modules, which leaves room for improvements. Recent, a series of LoRA variants \cite{Zhang2023AdaptiveBA,Ding2023SparseLA,Hu2023StructureAwareLA,zhang2024autolora,liu2024alora} propose different approaches to allocate or prune the uniform distributed LoRA ranks adaptively on the given task and LLM backbone, achieving performance boosts on the downstream tasks. However, the existing methods rely on un-reliable and in-explainable LoRA rank importance measurements, thus they still can not fully exploit the potential of LoRA fine-tuning. 

To address the above issues, we propose the novel \underline{Shap}ley Value inspired \underline{Lo}w-\underline{R}ank \underline{A}daptation framework (ShapLoRA). To address the shortcomings of the sensitivity-based importance measure, we propose to put this measure under different coalitions in the collaborative games among LoRA ranks, inspired by game theory based attribution method Shapley Value \cite{lundberg2017unified}. This results in a novel and more explainable importance measure, which is named as Shapley sensitivity, to pay homage to Shapley Value. In addition, we optimize the workflow of the existing works by: (a) conducting LoRA rank pruning once based on the Shapley sensitivity scores on a separate validation set instead of the training set. (b) Setting up the allocating-retraining procedures for fair comparisons among different PEFT methods.

\begin{figure*}
\centering
\includegraphics[width=0.8\textwidth]{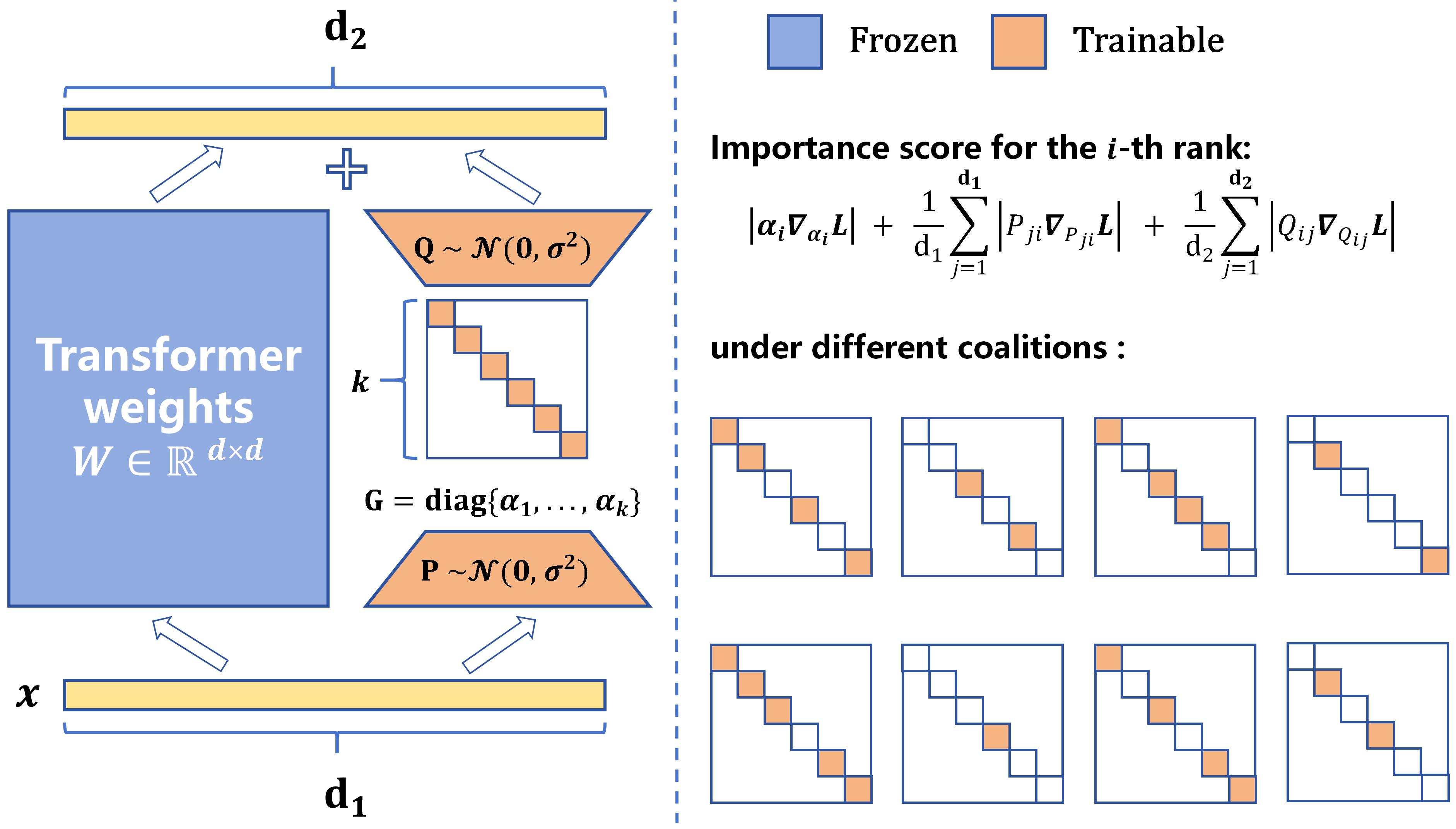}
\caption{Schematic illustration of our ShapLoRA framework. (\textbf{Left}): ShapLoRA follows LoRA and AdaLoRA to update the weight matrix $W$ by fine-tuning the low-rank incremental matrix $\Delta W = PGQ$ with a SVD formation. And LoRA rank $i$ is represented by a tuple ($G_{i,i}$, $P_{:,i}$, $Q_{i, :}$) of parameters. (\textbf{Right}): ShapLoRA prunes the less important LoRA rank $k$ by fixing $G_{k,k}$ to zero. Here the importance score $\text{IPT(k)}$ for each LoRA rank $k$ is calculated as the average sensitivity score of the parameters in rank $k$. Different from previous works, $\text{IPT(k)}$ is calculated under a random subset of coalitions in which each LoRA rank is considered as a player in a collaborative game. A coalition is formulated by randomly masking some of the elements of $G$ to zeros.   }
\label{fig:architecture}
\vspace{-1.5em}

\end{figure*}

We conduct comprehensive experiments across diverse challenging tasks, including question-answering tasks, math problem solving tasks, and a collection of natural language processing tasks. Our approach consistently surpasses strong parameter-efficient fine-tuning (PEFT) baselines, including state-of-the-art LoRA variants, under comparable trainable parameter budgets. The experimental results also demonstrate the general applicability of our ShapLoRA framework through training and inference efficiency analysis. Our contributions are summarized as follows: 
\begin{itemize}[left = 0em]
\item we propose ShapLoRA, a novel framework designed to adaptively adjust the LoRA rank allocation on the LLM backbone, resulting in improved fine-tuning quality. 

\item The core of our ShapLoRA framework is the our proposed Shapley sensitivity, an LoRA rank importance measure which combines the gradient-based sensitivity measure and game theory. 

\item Through comprehensive empirical evaluations and in-depth analysis, we demonstrate that our proposed ShapLoRA framework delivers high-performance LoRA rank allocation settings that consistently surpass baseline methods under equivalent tunable parameter budgets. 

\end{itemize}

\section{Related works}

Due to limited length, we put more literature reviews on parameter-efficient fine-tuning to Appendix \ref{sec:appendix_related_works}.

\subsection{The LoRA method and its variants}

The third line of works, which our work belongs to, address the issue of allocating LoRA ranks among the linear modules across different layers of the Transformer-based LLM backbones. AdaLoRA \cite{Zhang2023AdaptiveBA} expresses the low-rank multiplication of LoRA in the form of singular value decomposition (SVD), and it identifies the most important ranks by a sensitivity-based importance score. SoRA \cite{Ding2023SparseLA} prunes abundant LoRA ranks by imposing a $l_{0}$ norm and optimizing with proximal gradient descent. SaLoRA \cite{Hu2023StructureAwareLA} prunes the LoRA ranks via the Lagrange multiplier method. ALoRA \cite{liu2024alora} address the importance estimation of LoRA ranks via an noval heuristic measure. AutoLoRA \cite{zhang2024autolora} proposes to associate each LoRA rank with a selection variable, and updates these variables during the fine-tuning procedure. Then the pruning of LoRA ranks relies on the magnitude of these learned selection variables. AutoLoRA applies the idea of differentiable architecture search \cite{zhang2024autolora} into LoRA rank allocation, and these selection variables are referred to as the architectural parameters by \cite{Liu2019DARTSDA}. Despite these recent efforts, we believe issues still need to be investigated for LoRA rank allocation: the existing works replies on flawed importance estimations which can not reliably reflect the contribution of each LoRA rank. Our work complements the existing literature by proposing a novel LoRA rank importance measurement.

\section{Background}

\noindent \textbf{Transformer model} \quad Currently, the most widely used open-sourced large language models adopt the stacked Transformer architecture \cite{Vaswani2017AttentionIA}. The transformer block is primarily constructed using two key submodules: a multi-head self-attention (MHA) layer and a fully connected feed-forward (FFN) layer. Denote the input sequence's length as $l$, the hidden states' dimension as $d_{model}$, and the dimension at the FFN module as $d_{ffn}$. The MHA is given as follows:\footnote{We omit the multi-head setting for simplicity of illustrations. }
\begin{equation}
\text{softmax} \left( \dfrac{ Q K }{\sqrt{d_{model}} } \right) V,
\label{eq:self_attn_1}
\end{equation}
where $Q = xW^{Q}$, $K = xW^{K}$, $V = xW^{V}$, $x \in \mathbf{R}^{l\times d_{model} }$ is the input tensor. $W^{Q}, W^{K}, W^{V} \in \mathbf{R}^{d_{model} \times d_{model} }$ are the query, key, and value projection layers (denoted as the Query, Key, and Value modules, or the Q, K, V modules). The FFN module consists of linear transformations and an activation function $g^{ffn}$ such as ReLU or GELU \cite{Hendrycks2016GaussianEL}. Take the FFN module in the LlaMA-2 models \cite{Touvron2023Llama2O} as example:
\begin{equation}
(g^{ffn}( G ) * U ) W^{D}, 
\label{eq:ffn_1}
\end{equation}
where $G = xW^{G}$, $U = xW^{U}$, $W^{G}, W^{U} \in \mathbf{R}^{ d_{model} \times d_{ffn} } $ (denoted as Gate and Up module, or the G and U modules). The number of linear modules in a Transformer block is denoted as $N_{mod} > 0$. Thus, in LlaMA-2, $N_{mod} = 7$.

\noindent \textbf{Low-rank adaptation (LoRA)} \quad  For any Transformer's linear module $m \in \text{\{Q, K, V, O, G, U, D\}}$ on the $l$-th ($1 \leq l \leq L$, $L> 0$ denoting the number of Transformer layers in the LLM), AdaLoRA \cite{Zhang2023AdaptiveBA} formulates LoRA in a singular value decomposition (SVD) format:
\begin{equation}
x^{'} = xW^{l, m} + b^{l, m} + x P^{l, m} \Lambda^{l, m} Q^{l, m}, 
\label{eq:lora_eq_2}
\end{equation}
where $P^{l, m} \in \mathbb{R}^{d_{1} \times r }$ and $Q^{l, m} \in \mathbb{R}^{ r \times d_{2} }$ are the left/right singular vectors, $\Lambda^{l, m} \in \mathbb{R}^{ r \times r }$ is the diagonal matrix contains the singular values $\{ \lambda^{l, m}_i \}_{i = 1}^{r}$. These three matrices contain the tunable parameters for the LoRA module. $\Lambda^{l, m}$ is initialized with zero while $P^{l, m}$ and $Q^{l, m}$ adopt a random Gaussian initialization to ensure $P^{l, m} \Lambda^{l, m} Q^{l, m}$ is a zero matrix at the beginning of training. Under Equation \ref{eq:lora_eq_2}, the $i$-th rank of LoRA $(l, m)$ contains the $i$-th singular value and vectors, and is denoted as $\mathcal{G}_{i}^{l, m} = (\lambda_{i}^{l, m}, P^{l, m}_{*, i}, Q^{l, m}_{i, *})$. Note that during the training process of AdaLoRA, LoRA rank pruning is conducted by setting some of the singular values to zero step-by-step.

\section{Method}

\subsection{Motivation}

We now reflect on the previous representative works on LoRA rank allocation. AdaLoRA \cite{Zhang2023AdaptiveBA} first consider re-arrage the rank distributions of LoRA modules on the LLM backbone. It achieve this objective by first initialize all the LoRA modules with a large number of ranks, and prune the less important ranks gradually along with the training procedure. AdaLoRA utilize a sensitivity based importance score \cite{michel2019sixteen}, 
\begin{equation}
\text{ipt}(w) = \| w \nabla_{w} \mathcal{L} \|
\end{equation}
which measures how much the training loss will change if the LoRA parameters change. However, \cite{zhang2022platon} pointed out that this importance measure is unreliable, since it only considers how one parameter change affects the model under the hypothesis that no other parameter changes occur, and have not consider its importance under different model statuses. 

AutoLoRA \cite{zhang2024autolora} builds upon the methodology of differentable neural architecture search and bi-level optimization \cite{Liu2019DARTSDA}. It considers each LoRA rank as a neural network operation and assigns a learnable architectural parameter. Its objective is to select the best LoRA architecture, which relies on the learned architectural parameters' values as the importance scores. SoRA \cite{Ding2023SparseLA} and SaLoRA \cite{Hu2023StructureAwareLA} are similar to AutoLoRA except that the architectural parameters are learned with a normal optimization procedure on the training set \cite{Bi2020GOLDNASGO}. The LoRA ranks with higher architectural weights are kept while others are pruned. However, as pointed out by \cite{chen2020stabilizing}, the architectural parameters can not reliably reflect the quality or importance of the LoRA ranks.

To summarize, to further enhance the performance of LoRA under rank allocation, we need to find an importance measure that is reliable and explainable. Thus, we draw inspiration from Shapley Value \cite{lundberg2017unified}. Shapley Value considers the collaborations among a group of players. In the context of this work, a player in the game is a LoRA rank. For player $k$ in the game, denotes all the other permutations of the other players as $\mathcal{S}_{k}$, then player $k$'s Shapley Value $\Phi_{k}$ is defined as:
\begin{equation}
\Phi_{m} = \dfrac{1}{|\mathcal{S}_{k}|} \sum_{A \in \mathcal{S}_{k} } V(A \cup \{ k \} ) - V( A ),
\label{eq:lora_shapley}
\end{equation}
where $V()$ denotes the performance metric for any coalition $A \in \mathcal{S}_{k}$. Intuitively, the calculation of the player $k$'s Shapley Value involves evaluating its contributions across different coalitions, that is, evaluating this player under different scenarios. However, Shapley Value is extremely time consuming, prohibiting its applications in deep learning. However, the core idea of Shapley Value could be the guide for our work.

\subsection{Shapley Sensitivity}

Now we introduce the core our ShapLoRA method, Shapley Sensitivity. This method combines the core idea of Shapley Value with the gradient-based sensitivity method \cite{michel2019sixteen}. In order to evaluate the LoRA rank $\mathcal{G}_{i}^{l, m}$ from LoRA $(l, m)$ in the style of Shapley Value, we put it under different coalitions by randomly masking the singular values of LoRA modules. For the LoRA rank $\mathcal{G}_{i}^{l, m}$, all the other LLM LoRA ranks' permutations are denoted as $\mathcal{S}_{i}^{l, m}$. If one wants to evaluate the LLM's performance under a permutation $S \in \mathcal{S}_{i}^{l, m}$, we need to exclude all the LoRA ranks not in $S$ to zeros:
\begin{equation}
\lambda_{ i^{'} }^{ l^{'} , m^{'} } = \begin{cases}
\lambda_{ i^{'} }^{ l^{'} , m^{'} } & \text{if } \mathcal{G}_{ i^{'} }^{l^{'}, m^{'}} \in  S \\
0 & \text{if } \mathcal{G}_{ i^{'} }^{l^{'}, m^{'}} \notin S
\end{cases}.
\end{equation}
Then, we calculate the importance score $\text{SAN}(\mathcal{G}_{i}^{l, m} | S)$ of LoRA rank $\mathcal{G}_{i}^{l, m}$ under permutation $S$:
\begin{equation}
\begin{aligned}
\text{SAN}(\mathcal{G}_{i}^{l, m} | S) & = \text{ipt}(\lambda_{i}^{l, m})  + \frac{1}{d_1} \sum_{j=1}^{d_1} \text{ipt}( P_{j, i}^{l, m} )   + \frac{1}{d_2} \sum_{j=1}^{d_2} \text{ipt}( Q_{i, j}^{l, m} ), 
\end{aligned}
\end{equation}
where $\text{ipt}(w)$ denotes the gradient-based sensitivity score for a model parameter:
\begin{equation}
\text{ipt}(w) = \| w \nabla_{w} \mathcal{L} \|,
\end{equation}
in which $\mathcal{L}$ denotes the loss objective during fine-tuning. Then, the Shapley sensitivity $\text{SAN}(\mathcal{G}_{i}^{l, m})$ of LoRA rank $\mathcal{G}_{i}^{l, m}$ is given by:
\begin{equation}
\text{SAN}(\mathcal{G}_{i}^{l, m}) = \dfrac{1}{ | \mathcal{ S }_{i}^{l, m} | } \sum_{ S \in \mathcal{ S }_{i}^{l, m} } \text{SAN}(\mathcal{G}_{i}^{l, m} | S). 
\label{eq:san_calculation}
\end{equation}

Note that although more efficient than Shapley Value, the above equation for calculating Shapley sensitivity is still prohibitively computation-consuming. The next subsection will introduce an approximation to $\text{SAN}(\mathcal{G}_{i}^{l, m})$.

\subsection{Workflow of ShapLoRA}

We now describe the complete workflow of our ShapLoRA method during the rank allocation stage, which is based on our proposed Shapley sensitivity score. The training set for a downstream task is denoted as $\mathcal{D}_{train}$, and the validation set is denoted as $\mathcal{D}_{v}$. 

Initially, all the LoRA modules with equal ranks $r_{init}$ are installed on the LLM backbone. Thus the initial total number of ranks is $R^{init} = N_{mod} * L * r_{init}$. And our targeted total number of LoRA ranks is $R^{target} > 0$. These LoRA parameters are fine-tuned for  $K_1 > 0$ epochs to ensure convergence. Then the LoRA ranks' Shapley sensitivity scores will be calculated on $\mathcal{D}_{v}$, and $R^{prune} = R^{init} - R^{target}$ ranks with lower scores will be pruned, and we obtain the rank allocation configurations. 

Since Equation \ref{eq:san_calculation} is impractical, we need to provide an approximation to strike a balance between efficiency and performance. Denote $\mathcal{S}_{all}$ as all the permutations of the LoRA ranks. Then a subset $\mathcal{S}_{sub}$ of size $N_3 > 0$ is drawn from $\mathcal{S}_{all}$. \footnote{Once a $S \in \mathcal{S}_{all}$ is drawn, we also draw its complementary $S^{'}$ to $\mathcal{S}_{sub}$. That is $S \cup S^{'}$ contains all the LoRA ranks. This ensures that all the LoRA ranks will be masked with equal times. } And the the Shapley sensitivity scores for any LoRA rank $\mathcal{G}_{i}^{l, m}$ is given by:
\begin{equation}
\text{SAN}(\mathcal{G}_{i}^{l, m}) \approx \dfrac{ \sum_{ S \in \mathcal{S}_{sub} } \text{SAN}(\mathcal{G}_{i}^{l, m} | S) }{ | \mathcal{S}_{sub} | }.
\label{eq:san_approximate}
\end{equation}

\section{Experiments}

In this section, we conduct extensive experiments to evaluate our ShapLoRA method.

\subsection{Datasets and evaluation metrics}

We compare our approach to the baselines on a collection of challenging tasks: (a) five benchmark common-sense question-answering tasks, ARC-e and ARC-c \cite{clark2018think}, OBQA \cite{mihaylov2018can}, PIQA \cite{bisk2020piqa}, BoolQ \cite{clark2019boolq}. (b) two math reasoning tasks,  AQuA \cite{ling2017program} and  GSM8k \cite{cobbe2021training}. (c) MT-Bench \cite{2023arXiv230605685Z}, MMLU \cite{hendrycks2020measuring}, and BBH \cite{suzgun2022challenging}. Since these tasks provide no training data, we utilize the UltraChat \cite{ding2023enhancing} dataset for instruction tuning. (d) A collection of natural language processing (NLP) or natural language generation (NLG) tasks, including SST-2, RTE, QNLI from the GLUE benchmark \cite{Wang2018GLUEAM}, a conditional generation task E2E \cite{novikova-etal-2017-e2e}, and a SQL generation task WikiSQL \cite{Zhong2017Seq2SQLGS}.

Dataset introductions, statistics and evaluation metrics are introduced in Appendix \ref{sec:appendix_datasets}.

\subsection{Baselines}

We compare our ShapLoRA framework with the current SOTA PEFT baseline methods. 

\noindent\textbf{LoRA and its variants} \ we consider the following LoRA variants as baselines: (a) the original LoRA \cite{hu2021lora}; (b) AdaLoRA \cite{Zhang2023AdaptiveBA}, which adaptively adjust the LoRA parameters among different Transformer modules. (c) AutoLoRA \cite{zhang2024autolora}, which utilize the bi-level optimization method \cite{Liu2019DARTSDA} to learn the LoRA ranks' importance scores. (d) MOELoRA \cite{liu2023moelora}, which considers each LoRA module as a mixture of single-rank LoRA experts. (e) DoRA \cite{liu2024dora}.

\noindent\textbf{Other PEFT methods} \ We also consider the most recent PEFT methods: (a) Parallel-Adapter proposed by \citet{He2021TowardsAU}; (b) Learned-Adapter \cite{Zhang2023LearnedAA}. (c) P-tuning v2 \cite{Liu2021PTuningVP}. (d) IAPT \cite{zhu2024iapt}. (e) BitFit \cite{BenZaken2021BitFitSP}. (f) (IA)$^{3}$ \cite{Liu2022FewShotPF}, which multiplies learnable vectors to the hidden states in different modules of the Transformer layer. (g) SSP \cite{Hu2022SparseSS}.

The baselines are implemented using their open-sourced codes. We only adjust the hyper-parameters related to tunable parameter numbers to fairly compare the baseline methods and our ShapLoRA method. The hyper-parameter settings for the baselines are detailed in Appendix \ref{sec:appendix_exp_settings}.

\begin{table*}[t]
\centering
\resizebox{\textwidth}{!}{
\begin{tabular}{c|c|ccccccc|c}
\toprule\hline
\multirow{2}*{\textbf{Method}}   &   \textbf{Tunable}    &     \textbf{ARC-e}   &     \textbf{ARC-c}   &   \textbf{BoolQ}   &   \textbf{OBQA}  &  \textbf{PIQA}    &   \textbf{AQuA}   &    \textbf{GSM8k}    &   \multirow{2}*{\textbf{Avg.}}    \\ 
&  \textbf{Params}  &      \textbf{(acc)}   &   \textbf{(acc)}     &  \textbf{(acc)}   &   \textbf{(acc)}  &   \textbf{(acc)}  &   \textbf{(acc)}  &   \textbf{(acc)}    &     \\
\hline

\multicolumn{10}{c}{\textbf{\emph{Baselines}}}  \\
\hline

Parallel-Adapter  &    20.9M   &     72.4    &    54.2    &     75.3    &   76.3    &   69.8   &    45.6   &  56.4    &    64.3  \\
Learned-Adapter   &   21.1M   &     73.1     &    54.4  &    74.9    &  78.4   &    75.6   &   48.3   &    58.9     &     66.2   \\
\hline
P-tuning v2    &    21.0M      &    68.5   &     51.3    &  71.2    &  76.1    &  66.2    &   39.63    &  51.1   &    60.6   \\

IAPT    &    20.9M    &      75.1    &     54.7    &  77.8    &  79.2    &    77.3     &   43.6     &   55.8     &    66.2    \\
\hline
BitFit &    25.2M   &     72.3     &    54.1    &   76.4   &   77.2    &     76.6    &   41.8     &    51.7    &  64.3  \\
(IA)$^{3}$  &    23.1M   &       73.1     &   54.6      &    77.2    &  78.1    &   75.4     &   43.2   &   53.4     &   65.0   \\
SSP &   80.6M   &      75.2    &  57.6    &     79.6    &    79.5      &    79.7    &    45.9     &    61.8      &   68.5  \\
\hline

LoRA   &     22.5M     &     74.4    &    57.2   &    78.8   &    81.1   &    81.4    &     46.6     &  61.1     &   68.7   \\ 
AdaLoRA   &  23.2M     &    75.1    &    57.9   &    79.2   &    81.4   &    82.1    &  47.6    &   61.7     &   69.3   \\
AutoLoRA  &    23.1M   &    76.9   &   59.6  &   80.3  &   \underline{81.7}   &   82.5   &    47.9   &   61.3    &   70.0  \\
MOELoRA &  29.9M       &     77.5   &  \underline{60.2}   &     81.4     &    81.7    &     82.4    &     \underline{48.3}      &  62.3     &   70.5   \\
DoRA   &    22.6M     &    \underline{77.9}   &   59.8   &    \underline{81.7}    &    81.6   &    \underline{82.7}  &    47.9    &   \underline{62.6}      &  \underline{70.6}   \\
\hline

\multicolumn{10}{c}{\textbf{\emph{Our proposed methods}}}  \\
\hline

\rowcolor{yellow!25}
\textbf{ShapLoRA}  &  \textbf{22.8M}      &   \textbf{79.3}   &  \textbf{61.1}   &  \textbf{82.8}  &  \textbf{82.9}  &  \textbf{84.5}   &   \textbf{49.7}    &    \textbf{64.4}     &    \textbf{72.1}    \\

\hline
\bottomrule
\end{tabular}
}
\caption{\label{tab:results_main_1} The Overall comparison of different PEFT methods. The backbone model is LLaMA-3 8B. Bold and underline indicate the best and second-best results.}
 
\end{table*}

\subsection{Experiment Settings}
\label{subsec:experimental_settings}

\noindent\textbf{Computing infrastures} \quad We run all our experiments on NVIDIA A40 (48GB) GPUs. 

\noindent\textbf{Pretrained backbones} \quad The main experiments use the most recent open-sourced LLM, LlaMA-3 8B released by Meta \cite{dubey2024llama} as the pretrained backbone model. In the ablation studies, we will also use the distilled Qwen 3B models and distilled LlaMA 8B models from Deepseek R1 \cite{guo2025deepseek}. 

\noindent\textbf{Prediction heads and decoding} \quad When fine-tuning a LLM, we only consider the supervised fine-tuning (SFT) setting \cite{ouyang2022training}. After receiving a prompt or instruction, all the predictions are generated using the language modeling head (LM head). No additional prediction heads are installed for making categorical or numerical predictions. For decoding during inference, we use beam search with beam size 3.

\noindent\textbf{Hyper-parameter settings for ShapLoRA} \quad Our ShapLoRA divide the whole workflow into two stages, the LoRA rank allocation stage and the final fine-tuning stage. During the LoRA rank allocation stage, we add LoRA modules with rank $r_init = 16$ at each linear module of the Transformer block, and use the given task's training set to fine-tune the LoRA parameters till convergence. Then, each LoRA rank is evaluated by our Shapley sensitivity measure (Equation \ref{eq:san_approximate}). In Equation \ref{eq:san_approximate}, we set the size of $\mathcal{S}_{sub}$ to 90 by randomly masking each LoRA rank with a binomial distribution with parameter in \{ 0.1, 0.2, 0.3, 0.4, 0.5, 0.6, 0.7, 0.8, 0.9 \} repeatedly for 5 times. Then, the LoRA ranks that received lower Shapley sensitivity scores are pruned to meet the targeted average LoRA rank budget $R^{target} = \frac{R^{init}}{2}$. 

During the final fine-tuning stage, all the LoRA modules are randomly initialized according to the allocation setting delivered by the previous stage. 

The settings for training on both stages are specified in Appendix \ref{sec:appendix_exp_settings}.

\noindent\textbf{Reproducibility} \quad We run each task under five different random seeds and report the median performance on the test set of each task. 

Due to limited length, other experimental settings for the baseline methods and the training procedure are put in Appendix \ref{sec:appendix_exp_settings}.

\subsection{Main results}

\noindent \textbf{Results for QA and math tasks.} \quad In this setup, We compare ShapLoRA with baseline PEFT methods by employing these methods in fine-tuning on a challenging downstream task. The experimental results on the five commonsense reasoning QA tasks and two math reasoning tasks are presented in Table \ref{tab:results_main_1}. We present the number of tunable parameters in the second column. Table \ref{tab:results_main_1} reveals that our ShapLoRA method outperforms the baseline methods across all seven tasks, with comparable tunable parameters. In particular, ShapLoRA outperforms the previous SOTA LoRA-based baselines like AdaLoRA, AutoLoRA, DoRA, and MOELoRA with comparable or less tunable parameters. These results demonstrate that our method excels at downstream task adaptation of large language models.

\noindent \textbf{Results for general-purpose instruction tuning.} \quad After fine-tuning the large language model (LLM) (pretrained version) on the UltraChat dataset \cite{ding2023enhancing} using our proposed ShapLoRA method as well as AutoLoRA and MOELoRA\footnote{Due to limited resources, we now only provide the results for our ShapLoRA method and two representative and strong baselines on these LLM evaluation benchmarks. We will provide results for more baselines in the updated version.}, we evaluate performance across three challenging benchmarks: MT-Bench \cite{2023arXiv230605685Z}, MMLU \cite{hendrycks2020measuring}, and BBH \cite{suzgun2022challenging}. For MT-Bench, we report the average GPT-4 evaluation score (denoted as gpt4-score), while comprehensive results are detailed in Table \ref{tab:results_instruction_tuning}. Consistent with earlier findings (Table \ref{tab:results_main_1}), ShapLoRA achieves superior performance compared to MOELoRA across all benchmarks. These results underscore ShapLoRA’s efficacy in improving instruction-tuning quality for pretrained versions of the LLMs, highlighting its potential as a robust alternative to existing parameter-efficient adaptation strategies.

\noindent \textbf{Results on the GLUE and NLG tasks} \quad The experimental results on the three classification tasks and two NLG tasks are presented in Table \ref{tab:results_nlp_nlg}. Table \ref{tab:results_nlp_nlg} agains proves that our ShapLoRA method outperforms the strong baseline methods across all five tasks.

\subsection{Ablation studies and further analysis}
\label{subsec:ablation_studies}

\noindent\textbf{Analysis of the training and inference efficiency} \quad We now use the BoolQ task to analyze how much additional memory and time our ShapLoRA requires compared to the MOELoRA and AutoLoRA methods. Table \ref{tab:results_training_efficiency_analysis} presents the peak memory cost (in GiB) and time cost till obtaining the final LoRA parameters (in hours). All the methods requires comparable memory costs since the majority of the memory cost is due to the LLM backbone. MOELoRA requires few training hours since it does not require a two-stage workflow. AutoLoRA requires to learn to selection variables at the first stage with the help of bi-level optimization, which is time-costing. And it need to re-train the LoRA parameters after pruning. Our method's workflow is similar to AutoLoRA, and it turns out that the calculation of our Shapley sensitivity scores does not lead to excessive time costs.  Considering its superior performance compared to the baselines, our method is practical since its training costs is within a reasonable range.

\begin{table*}[t]
\centering

\begin{minipage}[t]{0.5\textwidth}
\centering
\renewcommand\arraystretch{1.2}
\begin{tabular}{c|ccc}
\toprule\hline
\multirow{2}*{\textbf{Method}} & \textbf{MT-Bench} & \textbf{MMLU} & \textbf{BBH} \\
& \textbf{gpt4-score} & \textbf{acc} & \textbf{acc} \\
\hline
MOELoRA & 7.31 & 56.6 & 47.8 \\
AutoLoRA & 7.39 & 57.1 & 47.6 \\
\hline
\textbf{ShapLoRA} & \textbf{7.56} & \textbf{58.7} & \textbf{48.7} \\
\hline\bottomrule
\end{tabular}
\caption{General-purpose instruction tuning performance.}
\label{tab:results_instruction_tuning}
\end{minipage}
\hspace{-10pt} 
\begin{minipage}[t]{0.5\textwidth}
\centering
\renewcommand\arraystretch{1.05}
\begin{tabular}{c|cc}
\toprule\hline
\multirow{2}*{\textbf{Method}} & \textbf{Time cost} & \textbf{Memory cost} \\
& \textbf{(hours)} & \textbf{(GiB)} \\
\hline
MOELoRA & 2.1 & 18.2 \\
AutoLoRA & 5.3 & 18.6 \\
\hline
\textbf{ShapLoRA} & 4.8 & 17.9 \\
\hline\bottomrule
\end{tabular}
\caption{Time and memory cost on BoolQ.}
\label{tab:results_training_efficiency_analysis}
\end{minipage}

\vspace{-0.8em}
\end{table*}
The inference efficiency analysis is presented in Table \ref{tab:results_inference_efficiency_analysis} in Appendix \ref{sec:app_inference_efficiency}, which shows that the LoRA configurations obtained by ShapLoRA has the comparable decoding speed with baselines.

\noindent\textbf{Visualization and analysis of the ShapLoRA's importance score distributions and rank allocation settings} \quad 
We present the LoRA ranks' Shapley sensitivity scores (normalized per linear module) at the 8th, 16th, 24th, and 32nd layer of LlaMA-3 8B as a heatmap in Figure \ref{fig:boolq_llama3_8b__lora_importance_scores} and \ref{fig:piqa_llama3_8b__lora_importance_scores} corresponding to the BoolQ and PIQA task, respectively in Appendix \ref{sec:appendix_e_lora_visualize}. And we present the obtained LoRA rank configurations after pruning in Figure \ref{fig:lora_allocation_results}. We can observe that: 
\begin{itemize}[left = 0em]
\item Different downstream tasks delivers different rank allocation configurations in our ShapLoRA framework, demonstrating the task specificity of LoRA fine-tuning. 

\item However, similarity in LoRA rank allocations across different tasks can be observed. We can observe from Figure \ref{fig:lora_allocation_results} that more LoRA ranks are pruned from the Query and Key modules, while the Value module keeps most of the LoRA ranks. 

\item The LoRA importance distributions across different modules and Transformer layers are different. Intuitively, different modules at Transformer layers play different roles, thus requiring different quantities of LoRA parameters.

\end{itemize}

\noindent\textbf{Ablation on the ShapLoRA framework} \quad In order to further demonstrate the effectiveness of ShapLoRA, and the appropriateness of its workflow designs, we now conduct a series of ablation studies on our workflow. We now consider the following variants of the ShapLoRA framework: (a) ShapLoRA-1 substitutes our proposed Shapley sensitivity score to the original sensitivity score \cite{michel2019sixteen}. That is, ShapLoRA-1 is equivalent to AdaLoRA with just one pruning step. (b) ShapLoRA-2 substitutes our proposed Shapley sensitivity score to the magnitude-based score \cite{guptaglobal}. (c) ShapLoRA-3 considers a pruning schedule similar to AdaLoRA, that is, pruning 4 LoRA ranks once every 25 steps. (d) ShapLoRA-4 considers calculating the Shapley sensitivity score on the training set. (e) ShapLoRA-5 reduces the times of random masking to 18 times. (f) ShapLoRA-6 increases the times of random masking to 900 times. 

The ablation experiments on the BoolQ, PIQA, and MMLU are presented in Table \ref{tab:ablations}. The results show that ShapLoRA under the default settings (as in Table \ref{tab:results_main_1}) outperforms or performs comparable to the six variants. In addition: (a) Comparing ShapLoRA to ShapLoRA-1 and ShapLoRA-2 demonstrates the effectiveness of our Shapley sensitivity score in identifying important LoRA ranks. (b) ShapLoRA-3 performs comparable to ShapLoRA, showing that a fine-grained pruning schedule is not necessary under our framework. (c) ShapLoRA-4 performs slightly worse than ShapLoRA, showing that it is better to determine which LoRA ranks to prune on a separate validation set to avoid receiving biased estimations of importance on the overfitted training data. We believe that this is one of the major reason why AutoLoRA outpeforms AdaLoRA. (d) Comparing ShapLoRA with ShapLoRA-5 and ShapLoRA-6 demonstrates that the random masking setting for calculating the Shapley sensitivity strikes a good balance between performance and efficiency.

\begin{table}[th!]
\centering
\begin{tabular}{c|ccc}
\toprule\hline
\multirow{2}*{\textbf{Method}}    &     \textbf{BoolQ}     &   \textbf{PIQA}   &    \textbf{MMLU}  \\ 

&    \textbf{(acc)}  &   \textbf{(acc)}   &   \textbf{(acc)}  \\
\hline
ShapLoRA   &    82.8  &    84.5   &      58.7     \\
\hline
ShapLoRA-1   &    81.8   &    83.6    &     57.9     \\
ShapLoRA-2   &    81.0   &    82.1    &     56.8     \\
ShapLoRA-3   &    82.8   &    84.6    &     58.6    \\
ShapLoRA-4   &    82.1   &    83.7    &     57.5     \\
ShapLoRA-5   &    82.2   &    83.3    &     57.6     \\
ShapLoRA-6   &    82.8   &    84.5    &     58.8         \\
\hline\bottomrule
\end{tabular}

\caption{\label{tab:ablations} The comparison of ShapLoRA's variants on the BoolQ, PIQA, and MMLU tasks. }

\end{table}

\begin{figure}[!ht]
\centering

\begin{subfigure}{0.45\textwidth}
    \centering
    \includegraphics[width=\linewidth]{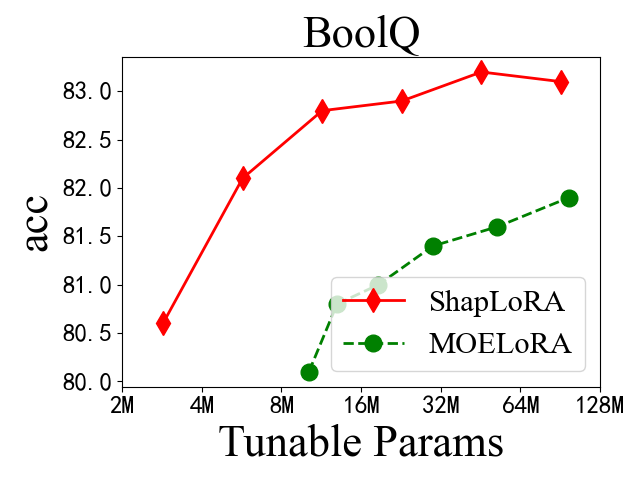}
    \caption{BoolQ}
    \label{subfig:BoolQ_different_tunable_paras}
\end{subfigure}
\hspace{4pt}
\begin{subfigure}{0.45\textwidth}
    \centering
    \includegraphics[width=\linewidth]{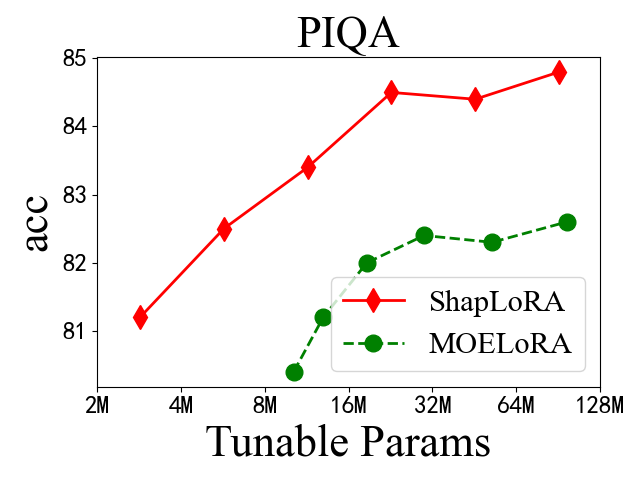}
    \caption{PIQA}
    \label{subfig:PIQA_different_tunable_paras}
\end{subfigure}

\caption{Performances under different tunable parameter budgets. The $x$-axis represents the number of tunable parameters, and the $y$-axis represents the performance score.}
\label{fig:different_tunable_paras}
\end{figure}

\noindent\textbf{Comparisons under different budgets of tunable parameters} \quad In our main experiments (Table \ref{tab:results_main_1}), we set the targeted average LoRA ranks in the ShapLoRA setting to $r_{target} = 8$. Now we change the budget of tunable parameters for ShapLoRA by modifying the $r_{target}$ to \{1, 2, 4, 16, 32\}. We also alter the MOELoRA method's tunable parameter numbers accordingly. The experimental results on the BoolQ and PIQA tasks are presented in Figures \ref{subfig:BoolQ_different_tunable_paras} and \ref{subfig:PIQA_different_tunable_paras}. The results show that under different tunable parameter budgets, our ShapLoRA method (a) can consistently outperform the MOELoRA method, and (b) is more robust to decreases in tunable parameter numbers.

\noindent\textbf{On the stability of the Shapley sensitivity scoring method} \quad
On a given LLM backbone, we examine the stability of our Shapley sensitivity score, which depends on random masking. We run the scoring procedure under three different random seeds and compare the resulting LoRA importance scores. The similarity between each pair of runs is measured using Spearman rank correlation. The three runs are not used in any earlier experiments.

The pairwise correlations are all close to one: the correlations involving Seed1 are 1.00 with itself, 0.99 with Seed2, and 0.96 with Seed3; the correlation between Seed2 and Seed~3 is 0.99. These high correlations show that the importance scores of the LoRA ranks remain stable across different random seeds, demonstrating that the Shapley sensitivity scoring used in ShapLoRA is robust to randomness in the masking process.

\noindent\textbf{Ablation on the pretrained backbones} \quad Our main experiments are conducted on the LlaMA-3 8B model. To demonstrate the wide applicability of our method, we now conduct experiments on the distilled Llama-3 8B and Qwen2.5 3B from Deepseek R1 \cite{guo2025deepseek}. The results are reported in Table \ref{tab:results_different_backbones} in Appendix \ref{sec:appendix_more_results}. We can see that on these two backbones, our method can also outperform the baseline methods.

\section{Conclusion}

In this work, we introduced the ShapLoRA framework to enhance the efficiency of LoRA fine-tuning. To address the drawbacks in the recents works on LoRA rank allocation, we propose Shapley sensitivity, a novel LoRA rank importance measurement which combines the gradient-based sensitivity and the idea of game-theory-based attribution methods.  We have conducted experiments on various challenging tasks, and the experimental results demonstrate that our ShapLoRA method can outperform the recent baselines with comparable tunable parameters.

\bibliography{reference}

\appendix

\section{Appendix: more related works}
\label{sec:appendix_related_works}

Since LoRA is the most popular PEFT method in the era of large language models, many works are devoted to improving upon LoRA. There are three representative lines of works. The first line of works is to apply LoRA to the quantized LLM backbone \cite{2023arXiv230514314D}. These works are important since they allow the democratization of LLM usage. QLoRA proposes a novel quantization method for LLMs, and provide extensive experiments on fine-tuning the quantized LLMs with LoRA. LowRA \cite{zhou2025lowra} is introduced, the first framework to enable LoRA fine-tuning below 2 bits per parameter with minimal performance loss, and highlights the potential of ultra-low-bit LoRA fine-tuning for resource-constrained environments. The second line of research includes a series of works \cite{chen2024llava,liu2023moelora,yang2024moral,dou2023loramoe,gou2023mixture} that looks into combining Mixture-of-Experts (MoE) \cite{shazeer2017outrageously,jacobs1991adaptive} and LoRA. MOELoRA \cite{liu2023moelora} proves that fine-tuning LoRA modules with a MOE router enables the LLMs to perform well in a multi-task learning setting. LoRAMoE \cite{dou2023loramoe} integrates LoRAs using a router network to alleviate world
knowledge forgetting after instruction tuning. MoCLE \cite{gou2023mixture} proposes a MoE architecture to activate task-customized model parameters based on instruction clusters.

\section{Appendix for the datsets and evaluation metrics}
\label{sec:appendix_datasets}

\subsection{Commonsense reasoning tasks}

\noindent\textbf{BoolQ} The BoolQ dataset, introduced by \cite{clark2019boolq}, is a benchmark dataset designed for training and evaluating models on the task of reading comprehension, specifically for answering yes/no questions. It comprises questions that are naturally occurring—sourced from real queries posed by people on various websites. Each question is paired with a corresponding passage from Wikipedia that provides the necessary context to answer the question. The dataset is notable for its diverse and challenging nature, featuring questions that require a deep understanding of the passage, inference, and sometimes common sense reasoning.

\noindent\textbf{OpenBookQA} The OpenBookQA \cite{mihaylov2018can} dataset is a benchmark designed to evaluate the ability of AI systems to understand and reason with elementary-level science knowledge. Created by the Allen Institute for AI, it includes multiple-choice questions, each with four possible answers. The questions are based on a core set of science facts that are typically found in a student's "open book" of basic science knowledge. Unlike straightforward fact-recall questions, OpenBookQA challenges models to apply, analyze, and reason about the facts, often requiring external common-sense knowledge to arrive at the correct answer. 

\noindent\textbf{ARC} The AI2 Reasoning Challenge (ARC) dataset \cite{clark2018think}, developed by the Allen Institute for AI (AI2), is a benchmark for evaluating the ability of AI systems to perform complex reasoning over science questions. The dataset is composed of science exam questions spanning multiple grade levels from third grade to ninth grade, collected from various sources such as textbooks, standardized tests, and other educational materials. The questions are divided into an Easy Set (ARC-e) and a Challenge Set (ARC-c), with the latter containing questions that require more sophisticated reasoning and understanding of scientific concepts.

\noindent\textbf{PIQA} The Physical Interaction Question Answering (PIQA) dataset \cite{bisk2020piqa} is designed to evaluate a model's understanding of physical interactions and common-sense reasoning. Developed by the Allen Institute for AI, PIQA consists of multiple-choice questions that focus on everyday scenarios and the practical use of objects. Each question presents a short description of a physical task and provides two possible solutions, challenging the model to select the most plausible one based on general physical knowledge and intuitive reasoning.

\subsection{Math reasoning tasks}

\noindent\textbf{AQuA} The AQuA (Algebra Question Answering) dataset \cite{ling2017program} is a comprehensive collection of algebraic problems designed to evaluate and enhance the problem-solving abilities of AI systems. It includes a wide range of questions covering various algebraic concepts, from basic arithmetic to more complex equations and word problems. Each problem is meticulously curated to test the system's ability to understand, interpret, and solve algebraic expressions and equations.

\subsection{Natural language understanding tasks}

We experiment on three natural language understanding tasks from the GLUE \cite{Wang2018GLUEAM} benchmark: 
\begin{itemize}
\item The Stanford Sentiment Treebank (SST-2) is a widely used benchmark task in natural language processing (NLP) for binary sentiment classification. It consists of sentences extracted from movie reviews in the Rotten Tomatoes dataset, annotated to indicate whether the expressed sentiment is positive or negative. SST-2 simplifies the original Stanford Sentiment Treebank (SST), which included fine-grained sentiment labels, by focusing on a two-class classification problem. Each sentence is labeled at the phrase level, but the task is typically evaluated using sentence-level predictions. 

\item The Recognizing Textual Entailment (RTE) task, part of the GLUE (General Language Understanding Evaluation) benchmark, is a natural language inference challenge designed to evaluate a model’s ability to determine whether a given hypothesis can be logically inferred (entailed) from a premise. Framed as a binary classification problem, RTE requires models to predict "entailment" (if the hypothesis necessarily follows from the premise) or "not entailment" (if it does not). 

\item The Question Natural Language Inference (QNLI) task, part of the General Language Understanding Evaluation (GLUE) benchmark, is designed to evaluate a model's ability to determine the relationship between a question and a given sentence. Adapted from the Stanford Question Answering Dataset (SQuAD), QNLI reformulates question answering as a binary classification problem. Each instance pairs a question with a sentence from a context passage, and the task is to predict whether the sentence contains the correct answer to the question (labeled as "entailment") or not ("not\_entailment"). 

\end{itemize}

\subsection{Natural language generation tasks}

We experiment on three natural language generation tasks:
\begin{itemize}
\item The E2E benchmark \cite{novikova-etal-2017-e2e} dataset for training end-to-end, data-driven natural language generation systems in the restaurant domain. It asks a model to generate natural utterances based on a set of given key contents. This dataset has a 42061/4672/4693 train/dev/test split.  

\item WikiSQL \cite{Zhong2017Seq2SQLGS} consists of a corpus of 87,726 hand-annotated SQL query and natural language question pairs. These SQL queries are further split into training (61,297 examples), development (9,145 examples) and test sets (17,284 examples). It can be used for natural language inference tasks related to relational databases. In this work, we will ask the LLMs to generate SQL queries based on the given natural language questions. 

\end{itemize}

\noindent\textbf{GSM8k} The GSM8k dataset \cite{cobbe2021training}, also known as the Grade School Math 8k dataset, is a comprehensive collection designed for evaluating and training mathematical problem-solving abilities of machine learning models. Comprising 8,000 high-quality, diverse grade school math word problems, GSM8k serves as a benchmark for assessing the performance of models in understanding and solving arithmetic, algebraic, and logical reasoning challenges. Each problem in the dataset is meticulously curated to reflect real-world scenarios that students encounter in grade school, ensuring relevance and practicality.

\subsection{The MMLU benchmark}

Massive Multitask Language Understanding (MMLU) \cite{hendrycks2020measuring} is a new benchmark designed to measure knowledge acquired during pretraining by evaluating large language models exclusively in zero-shot and few-shot settings. This makes the benchmark more challenging and more similar to how we evaluate humans. The benchmark covers 57 subjects across STEM, the humanities, the social sciences, and more. It ranges in difficulty from an elementary level to an advanced professional level, and it tests both world knowledge and problem solving ability. Subjects range from traditional areas, such as mathematics and history, to more specialized areas like law and ethics.

\subsection{The BBH benchmark}

BIG-Bench Hard (BBH) \cite{suzgun2022challenging} is a subset of the BIG-Bench, a diverse evaluation suite for language models. BBH focuses on a suite of 23 challenging tasks from BIG-Bench that were found to be beyond the capabilities of current language models. These tasks are ones where prior language model evaluations did not outperform the average human-rater.

\subsection{The MT-Bench dataset}

The MT-Bench \cite{2023arXiv230605685Z} dataset is a widely used dataset for evaluating the quality of LLMs. It contains 80 questions. The LLMs generate responses for these questions, and human annotators or LLM annotators will judge the quality of these responses.

\subsection{Instruction tuning datasets}

Instruction tuning is an important method to improve the general capabilities of large language models \cite{ouyang2022training}. With the rise of large language models in the scale of 10B parameters or more,  like GPT-3, T5, PaLM, researchers have actively explored the few-shot or zero-shot capabilities of these models. \cite{Mishra2021CrossTaskGV} find that fine-tuning these LLMs on a large scale datasets containing hundreds of NLP tasks significantly improves the zero-shot performances on unseen tasks, establishing the scaling law of task numbers. The previous works like \cite{Wei2021FinetunedLM} and T0 \cite{Sanh2021MultitaskPT} establishes the instruction tuning datasets by transforming the traditional NLP tasks into a unified prompt format. Instruct-GPT \cite{ouyang2022training} conducts instruction tuning using the dataset constructed based the user queries from the OpenAI API users. Note that this work is also a seminal work for human feedback learning with reinforcement learning. However, the complete instruction tuning dataset from \cite{ouyang2022training} remains closed. With the launch of ChatGPT, \cite{alpaca} (Alpaca) constructs an instruction tuning dataset with diverse topics using the self-instruct techniques. 

For our experiment, we employ two general-purpose instruction tuning datasets:
\begin{itemize}

\item The UltraChat dataset \cite{ding2023enhancing} is a large-scale, synthetically generated conversational corpus designed to train and enhance AI-driven dialogue systems. Comprising over 1.5 million multi-turn dialogues and 56 million conversation turns, it leverages OpenAI's GPT-3.5-turbo to simulate diverse, human-like interactions across a wide array of topics, including daily life, technical discussions, and creative role-playing scenarios. Structured as user-assistant exchanges, the dataset emphasizes naturalness, coherence, and contextual depth, supporting the development of robust language models capable of handling nuanced conversations. Available in English, Chinese, and Japanese, UltraChat serves as a versatile resource for fine-tuning large language models (LLMs), advancing research in dialogue systems, and improving cross-lingual applications. Its synthetic nature allows scalability while maintaining high-quality, varied interactions, making it accessible via platforms like Hugging Face for researchers and developers aiming to push the boundaries of conversational AI.

\item Alpaca dataset \cite{alpaca}. Specifically, we employs its cleaned version\footnote{\url{https://huggingface.co/datasets/yahma/alpaca-cleaned}.}. This dataset comprises 51K instructions and demonstrations, and is suitable for instruction tuning. The cleaned version corrects multiple issues such as hallucinations, merged instructions, and empty outputs.
\end{itemize}

The detailed statistics of the above tasks' datasets are presented in Table \ref{tab:dataset_stats}.

\begin{table*}[th!]
\centering
\resizebox{0.86\textwidth}{!}{
\begin{tabular}{cccccc}
\hline
Datasets  &  \#train    &  \#dev   &   \#test   &     Type   &   Metrics  \\ 
\hline
\multicolumn{6}{c}{\textbf{\emph{Commonsense reasoning tasks}}}   \\
\hline
BoolQ  &     9427	  &  -  &   3270   &  Commonsense reasoning   &  acc    \\
OBQA    &  4957	   &   500	    &   500     &  Commonsense reasoning   &  acc      \\

ARC-e   &    2251	 &   570   &   2376    &  Commonsense reasoning   &  acc      \\
ARC-c   &    1119	 &  299	  &   1172    &  Commonsense reasoning   &  acc      \\

PIQA   &   16,000   &    2,000     &    3,000   &  Commonsense reasoning   &  acc \\

\hline
\multicolumn{6}{c}{\textbf{\emph{Math reasoning tasks}}}  \\
\hline
AQuA    &    97467	 &   254	 &   254   &   Math reasoning  &       acc          \\
GSM8K  &   7473   &   -	   &  1319   &   Math reasoning  &       acc    \\

\hline
\multicolumn{6}{c}{\textbf{\emph{Natural language understanding tasks}}}  \\
\hline

SST-2  &  66k  &   1k    &   0.8k   &   Sentiment classification  &  acc     \\
RTE &   2.5k   &   0.1k   &    0.1k  &    Natural language inference   &     acc    \\
QNLI &  104k   &    1k   &    5.4k  &     Natural language inference   &     acc  \\

E2E &   42k  &  4.6k   &  4.6k    &  Natural language generation   &  rouge   \\
WikiSQL  &   61k  &  9K  &  17K    &   SQL generation  &   acc  \\

\hline
\multicolumn{6}{c}{\textbf{\emph{Instruction tuning }}}  \\
\hline

UltraChat   &    56M  &    -    &  -   &  Instruction tuning  &  -     \\

Alpaca   &    50k  &    -    &  -   &  Instruction tuning  &  -     \\

\hline
\multicolumn{6}{c}{\textbf{\emph{LLM evaluation tasks}}}  \\
\hline

MT-Bench   &   -   &  -  &   80  &   Question answering  &  GPT-4 scores       \\

MMLU  &  -  &  -  &  14042      &   Question Answering    &   acc    \\

BBH  &  -  &  -  &  6,511   &     Question Answering    &    acc    \\

\hline
\end{tabular}}
\caption{\label{tab:dataset_stats}  The dataset statistics. }
\end{table*}

\subsection{Evaluation metrics/protocols}
\label{sec:appendix_evaluations}

For the commonsense reasoning and math reasoning tasks, since they usually come with a definite answer choice, we will directly consider the correctness of the final answers. Thus, we report accuracy (denoted as acc).

For evaluating the quality of instruction tuned LlaMA-2 7B on the MT-Bench, we follow the current common practice of utilizing GPT-4 as a unbiased reviewer \cite{2023arXiv230605685Z}. We generate model responses from a fine-tuned model with beam size 3 with the generation function in Huggingface Transformers \cite{wolf2020transformers}. Then we compare MOELoRA and ShapLoRA's answers with GPT-4. For each instruction in MT-Bench, GPT-4 \cite{gpt4} is asked to write a review for both answers from the two methods, and assigns a quantitative score on a scale of 10 to each response. The prompts of instructing GPT-4 for evaluation is presented in Appendix \ref{sec:appendix_gpt4_eval}.

\section{Prompt templates for GPT-4 evaluations}
\label{sec:appendix_gpt4_eval}
In this work, we utilize the powerful LLM GPT-4 \cite{gpt4} as the evaluator for comparing the instruction tuning quality. As a reviewer, GPT-4 will receive a query [query], two responses, [response1] and [response2], from two assistants. We will ask GPT-4 to write a review for each response, assessing the quality of the response, and then ask GPT-4 to assign a score on a scale of 10 to each response.

Template for prompt: 
\begin{verbatim} 
Task Introduction
you will be given a query, and two responses 
from two assistants, 
could you compare the two responses, 
and do the following: 
(1) write a concise review for each 
assistant's response, on how well the 
response answers the query, and whether 
it will be helpful to humans users, and any 
issues in the response;
(2) assigns a quantitative score on a scale 
of 10 to each response, reflecting 
your assessment of the two responses
Query: 
[query]
Response 1 from assistant 1: 
[response1]
Response 2 from assistant 2: 
[response2]
\end{verbatim}

\section{Appendix for Experimental settings}
\label{sec:appendix_exp_settings}

Here, we provide more details for experimental settings. 

\noindent\textbf{Hyper-parameters for the baseline PEFT methods} \quad For P-tuning V2, the number of prompt tokens at each layer is set to 160. For IAPT, the prompt's length is set to 8, the bottleneck dimension is set to 256, and the number of prompt layers is set to 32. For adapter-based methods, the bottleneck dimension is set to 40, and the adapter modules are added on the self-attention and feed-forward module. For LoRA and ALoRA, the initial rank at each module is set to 8. For AdaLoRA and AutoLoRA, the initial rank at each module is set to 16, and half of the rank budget is pruned during fine-tuning. We adjust the sparsity for SSP so that the number of tunable parameters is comparable with ShapLoRA and the other baselines. For BitFit, the bias vectors are first initialized with 16 dimensions, and then are projected to the dimensions that are aligned with the linear modules of the Transformer backbone. For (IA)$^{3}$, the product vectors are first initialized with 32 dimensions, and then are projected to the dimensions that are aligned with the linear modules of the Transformer backbone.

\noindent\textbf{Training settings for PEFT methods} \quad We use the HugginFace Transformers \cite{wolf-etal-2020-transformers} and PEFT \cite{peft} for implementing all the methods, and for training and making predictions. For fine-tuning LlaMA-3 8B model, the maximum sequence length is set to 2048. The maximum training epoch is set to 10. The batch size is set between 16 for task with less than 10k training set, and 128 otherwise. We use AdamW as the optimizer with a linear learning rate decay schedule and 6\% of the training steps for warm-up. The learning rate is set to 1e-4. The other hyper-parameters are kept the same with \cite{wolf-etal-2020-transformers}. In every 200 steps, the model is evaluated on the dev set. Patience is set to 10, that is, if the model does not achieve a lower development set loss for 10 evaluation runs, the training stops. The best checkpoint on the dev set is used to run predictions on the test set.

\section{Appendix: more experimental results}
\label{sec:appendix_more_results}

\noindent \textbf{Results on the NLP and NLG tasks} \quad Table \ref{tab:results_nlp_nlg} reports the experimental results on the SST-2, RTE, QNLI, E2E, and WikiSQL tasks.

\begin{table*}[th!]
\centering
\resizebox{0.62\textwidth}{!}{
\begin{tabular}{c|ccccc}
\hline
\multirow{2}*{\textbf{Method}}    &     \textbf{SST-2}   &    \textbf{RTE}   &   \textbf{QNLI}   &   \textbf{E2E}   &    \textbf{WikiSQL}    \\ 

 &   \textbf{(acc)}   &   \textbf{(acc)}     &  \textbf{(acc)}   &       \textbf{(rouge)}     &     \textbf{(acc)}    \\
\hline
\multicolumn{6}{c}{\textbf{\emph{Baselines}}}  \\
\hline

Housbly-Adapter   &     94.3   &  86.7  &  93.1    &    72.2    &    86.8    \\
BitFit  &      94.5    &   86.9    &  93.7     &    72.7     &   87.3   \\
P-tuning v2    &        93.8    &   84.5  &  92.2     &  71.3    &  86.4    \\

LoRA   &        95.2   &   87.2    &  94.1  &   73.5    &   87.3       \\
AdaLoRA   &    95.2   &  87.3   &  94.0    &    73.8    &   86.9   \\

AutoLoRA   &    95.5   &  87.6   &  93.9    &    73.6    &   87.5   \\

\hline
\multicolumn{6}{c}{\textbf{\emph{Our proposed methods}}}  \\
\hline

ShapLoRA   &      \textbf{96.1}  &  \textbf{89.0}   &  \textbf{95.1}   &    \textbf{74.8}    &       \textbf{88.5}     \\

\hline
\end{tabular}}

\caption{\label{tab:results_nlp_nlg} The Overall comparison of the three GLUE tasks and two natural language generation tasks. The backbone model is LlaMA-3 8B.  We report the median performance over five random seeds. The metric for each task is explained in Appendix \ref{sec:appendix_evaluations}.} 
\end{table*}

\noindent \textbf{Results on different LLM backbones} \quad Table \ref{tab:results_different_backbones} reports the experimental results with different LLM backbones. 

\begin{table}[th!]
\centering
\resizebox{0.42\textwidth}{!}{
\begin{tabular}{c|ccc}
\hline
\multirow{2}*{\textbf{Method}}    &     \textbf{BoolQ}     &   \textbf{PIQA}   &    \textbf{MMLU}     \\ 

&    \textbf{(acc)}  &   \textbf{(acc)}   &   \textbf{(acc)}     \\
\hline 

\multicolumn{4}{c}{\textbf{\emph{Results for Deepseek distilled Llama-3 8B}}}  \\
\hline
MOELoRA   &     83.2     &    85.8    &   59.4 \\
\hline
ShapLoRA   &    \textbf{84.1}     &   \textbf{86.4}    &   \textbf{60.2}   \\

\hline 
\multicolumn{4}{c}{\textbf{\emph{Results for Deepseek distilled Qwen2.5 3B }}}  \\
\hline
MOELoRA   &   78.8     &    82.3    &   56.2 \\
\hline
ShapLoRA   &    \textbf{80.7}     &   \textbf{83.5}    &   \textbf{57.1}       \\

\hline

\end{tabular}}
\caption{\label{tab:results_different_backbones} Results for different PEFT methods on the BoolQ, PIQA and MMLU benchmarks. The backbone LLMs are the distilled Llama-3 8B and Qwen2.5 3B from Deepseek R1 \cite{guo2025deepseek}.}
\end{table}

\section{Appendix: inference efficiency}
\label{sec:app_inference_efficiency}

To demonstrate the inference efficiency of our ShapLoRA method, we now compare the GPU memory and decoding speed of ShapLoRA, AutoLoRA, and MOELoRA under beam search with different beam sizes. We use the test sets of the experimented tasks for efficiency evaluation. In this experiment, LoRA parameters are not merged to the backbone to mimic the single-LLM multi-LoRA setting \cite{Chen2023PunicaML}. We present two metrics for measuring efficiency: (a) peak memory cost (in GiB). (b) tokens generated per second (tps). The results are presented in Table \ref{tab:results_inference_efficiency_analysis}.  
From Table \ref{tab:results_inference_efficiency_analysis}, under beam sizes 1 and 3, the ShapLoRA method has a comparable decoding speed and memory cost with the other baselines.

\begin{table}[th!]
\centering
\resizebox{0.48\textwidth}{!}{
\renewcommand\arraystretch{1.05}
\begin{tabular}{c|ccc}
\hline
\multirow{2}*{\textbf{Method}}   &    \multirow{2}*{\textbf{Beam size}}  &  \textbf{Speed }   &   \textbf{Memory cost }     \\ 
&     &    \textbf{(tps)}    &   \textbf{(MiB)}  \\

\hline

\multirow{ 2}{*}{ AutoLoRA }   &   1    &   36.7     &    13.8    \\
        &   3   &    30.4    &   15.4    \\

\hline
\multirow{ 2}{*}{ MOELoRA }   &   1   &   32.5    &  13.8   \\
    &   3   &    26.4    &  15.6    \\

\hline
\multirow{ 2}{*}{ ShapLoRA }   &   1   &     37.1  &   13.7     \\
    &   3   &   30.7   &   15.2    \\
    
\hline
\end{tabular}}
\caption{\label{tab:results_inference_efficiency_analysis} The memory and speed of different PEFT methods during inference.  }
\end{table}

\section{Appendix: Visualization of Shapley sensitivity importance scores}
\label{sec:appendix_e_lora_visualize}

\begin{figure*}[t]
\centering

\begin{subfigure}{0.86\textwidth}
    \centering
    \includegraphics[width=\linewidth]{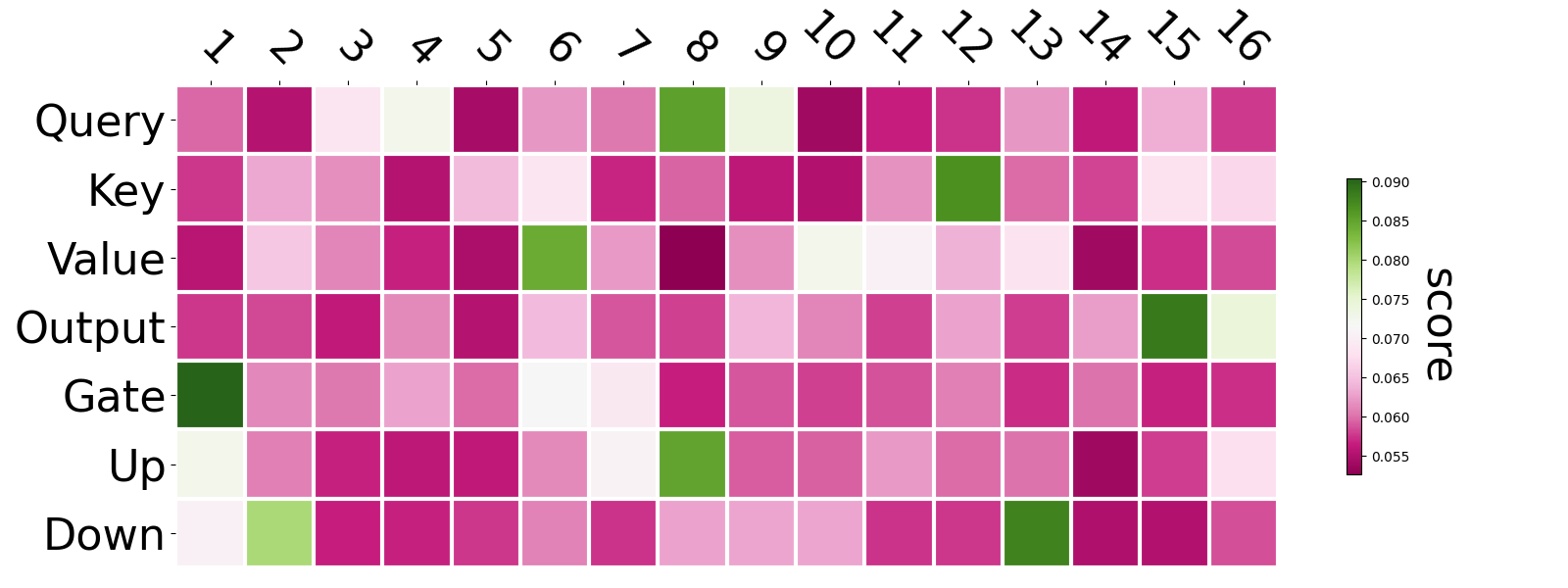}
    \caption{8th layer}
    \label{subfig:boolq_llama3_8b__layer_7_score_dist}
\end{subfigure}

\begin{subfigure}{0.86\textwidth}
    \centering
    \includegraphics[width=\linewidth]{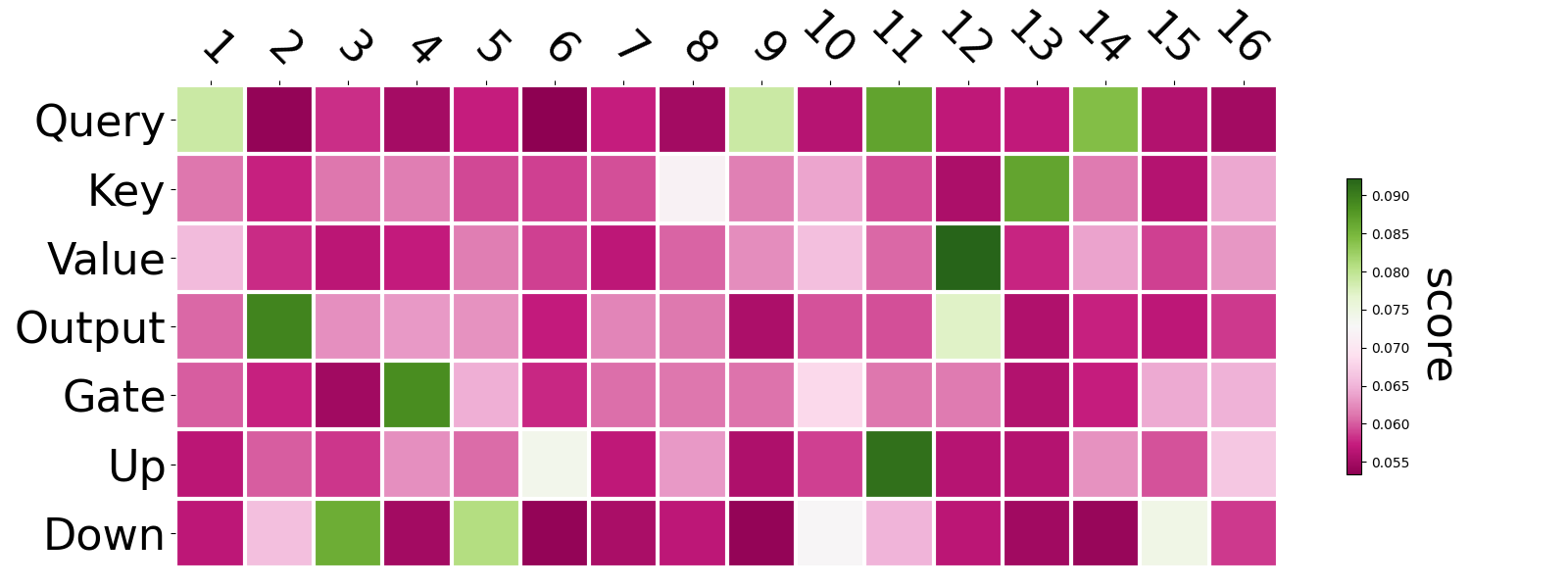}
    \caption{16th layer}
    \label{subfig:boolq_llama3_8b__layer_15_score_dist}
\end{subfigure}

\begin{subfigure}{0.86\textwidth}
    \centering
    \includegraphics[width=\linewidth]{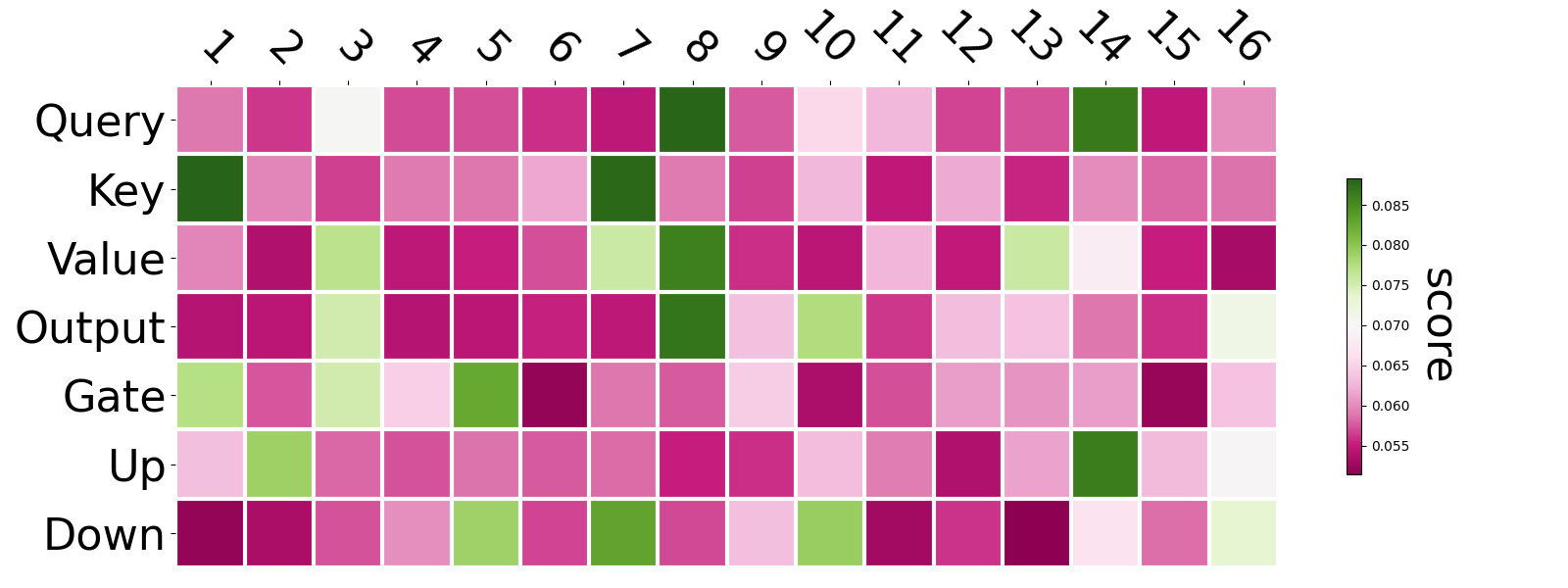}
    \caption{24th layer}
    \label{subfig:boolq_llama3_8b__layer_23_score_dist}
\end{subfigure}

\begin{subfigure}{0.86\textwidth}
    \centering
    \includegraphics[width=\linewidth]{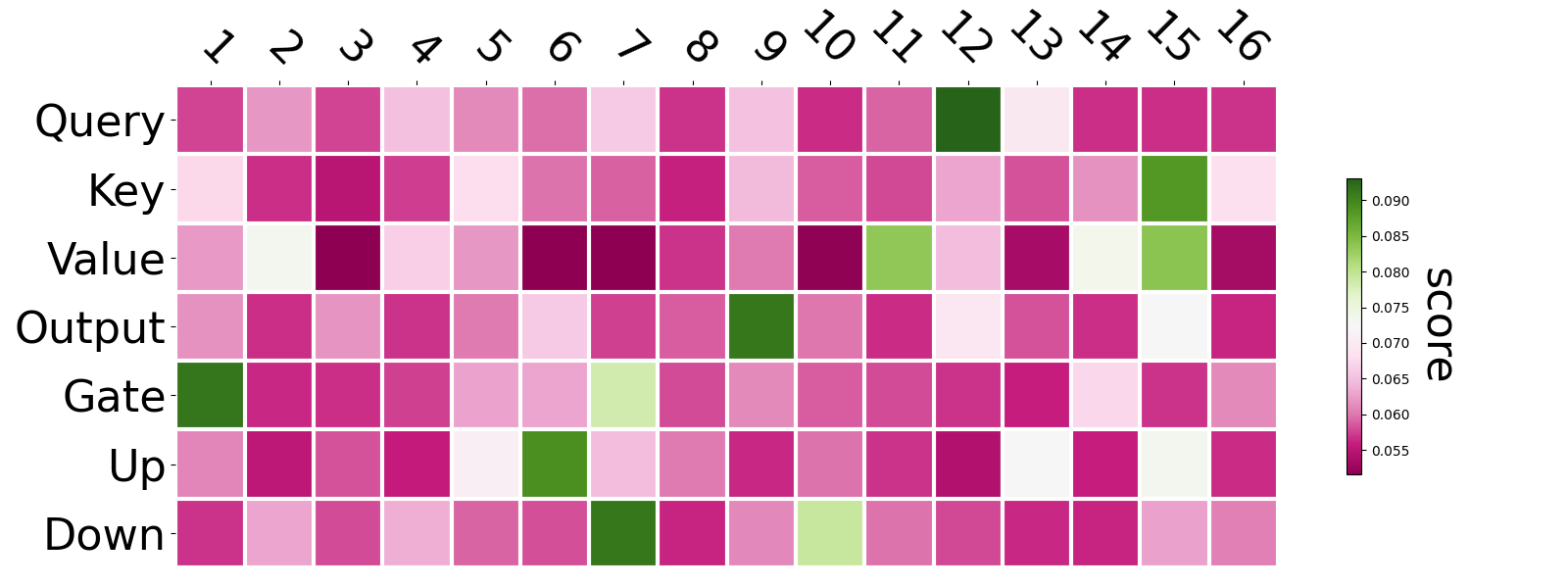}
    \caption{32nd layer}
    \label{subfig:boolq_llama3_8b__layer_31_score_dist}
\end{subfigure}

\caption{Shapley sensitivity importance scores of LoRA ranks on the 8th, 16th, 24th, and 32nd layers of LLaMA-3 8B after finetuning on BoolQ.}
\label{fig:boolq_llama3_8b__lora_importance_scores}
\end{figure*}

\begin{figure*}[t]
\centering

\begin{subfigure}{0.86\textwidth}
    \centering
    \includegraphics[width=\linewidth]{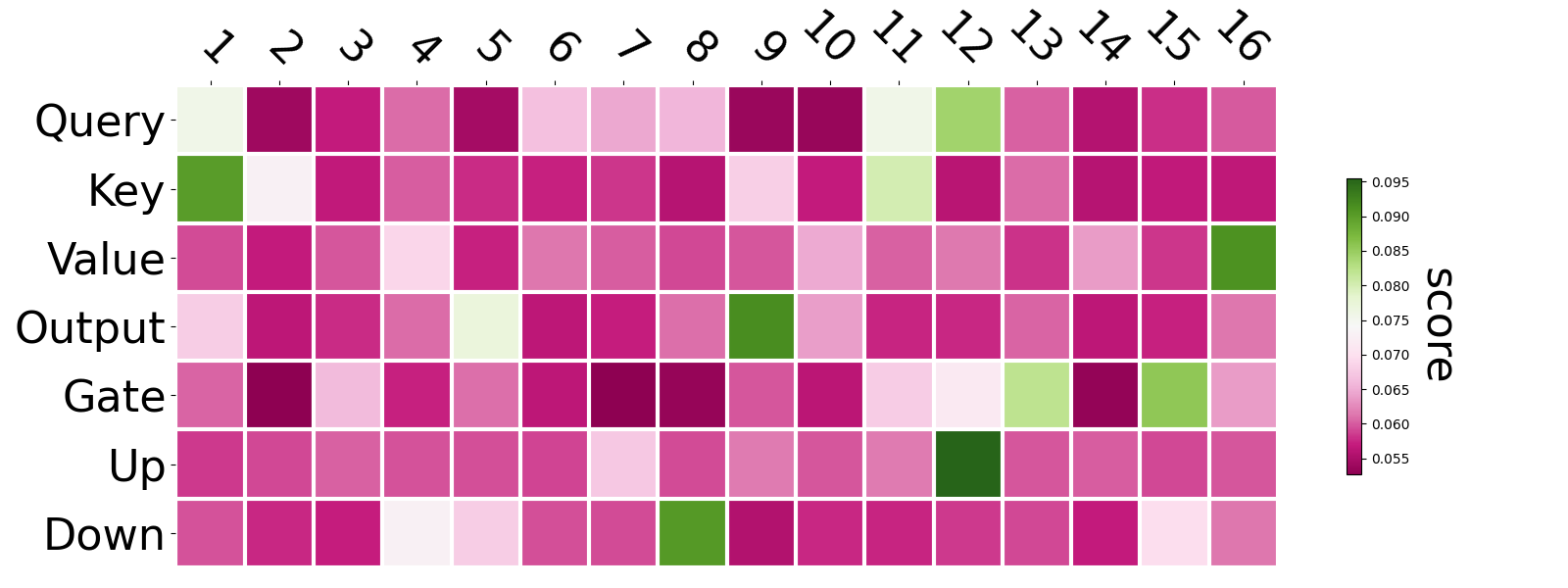}
    \caption{8th layer}
    \label{subfig:piqa_llama3_8b__layer_7_score_dist}
\end{subfigure}

\begin{subfigure}{0.86\textwidth}
    \centering
    \includegraphics[width=\linewidth]{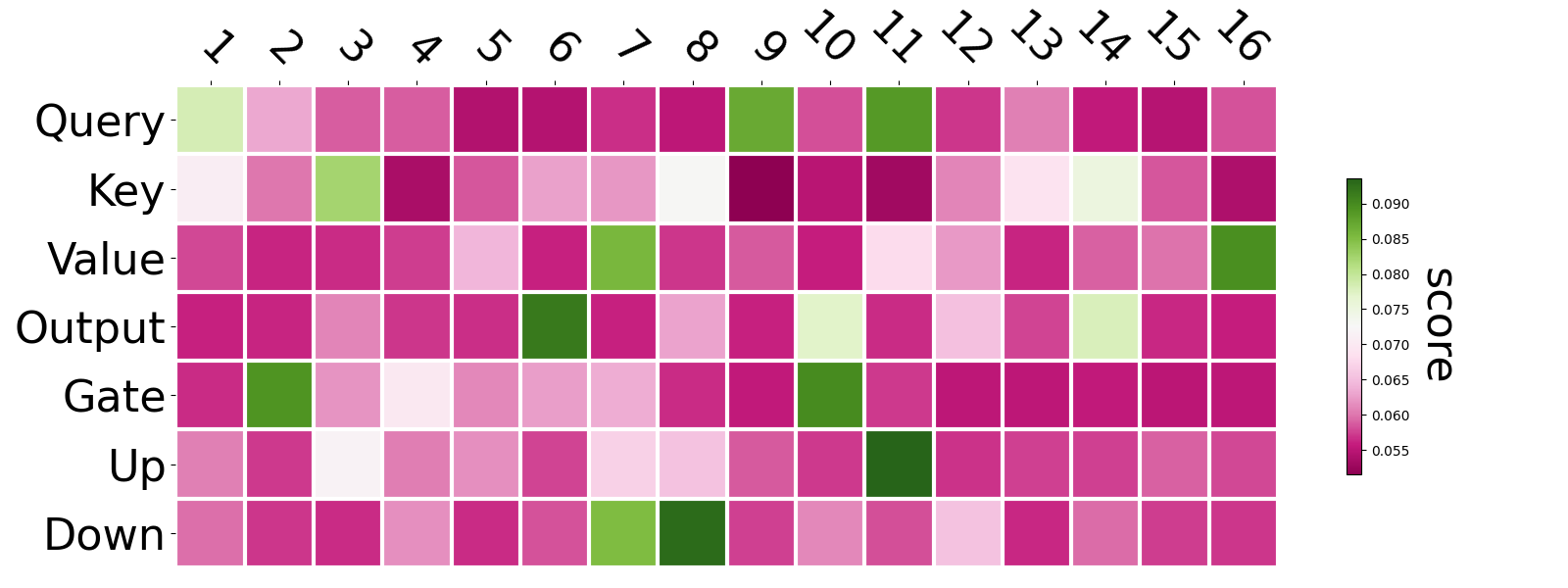}
    \caption{16th layer}
    \label{subfig:piqa_llama3_8b__layer_15_score_dist}
\end{subfigure}

\begin{subfigure}{0.86\textwidth}
    \centering
    \includegraphics[width=\linewidth]{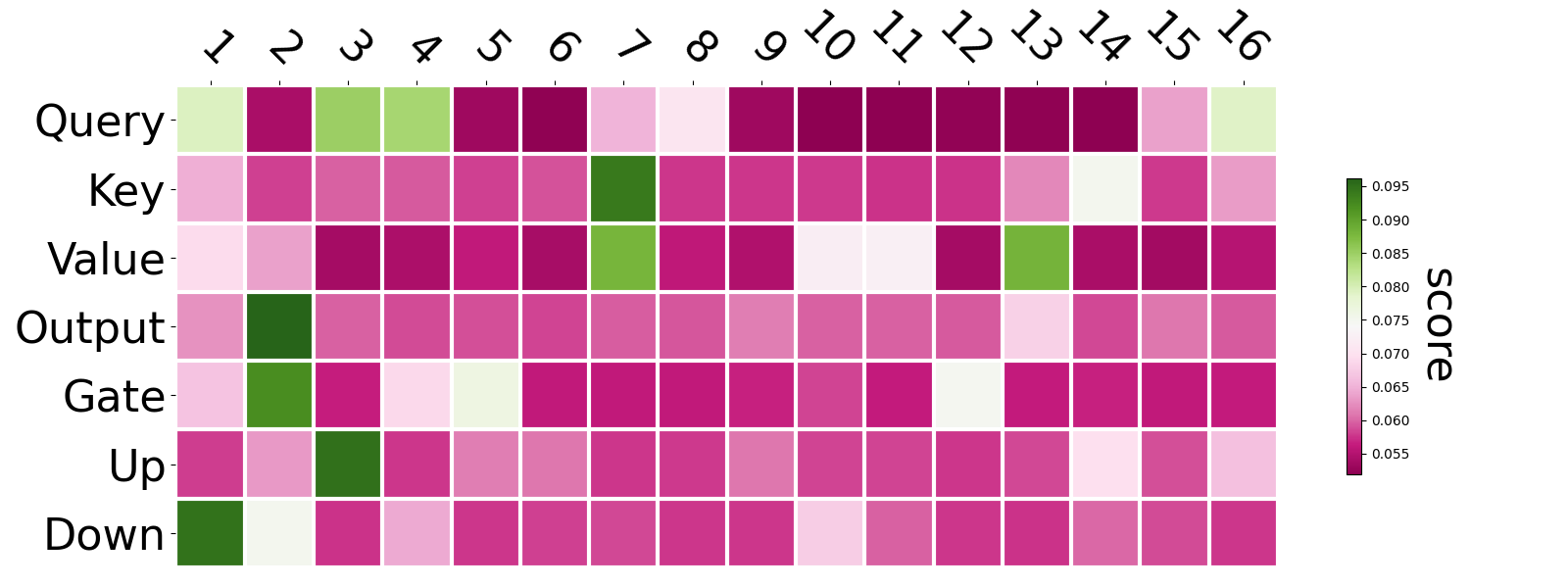}
    \caption{24th layer}
    \label{subfig:piqa_llama3_8b__layer_23_score_dist}
\end{subfigure}

\begin{subfigure}{0.86\textwidth}
    \centering
    \includegraphics[width=\linewidth]{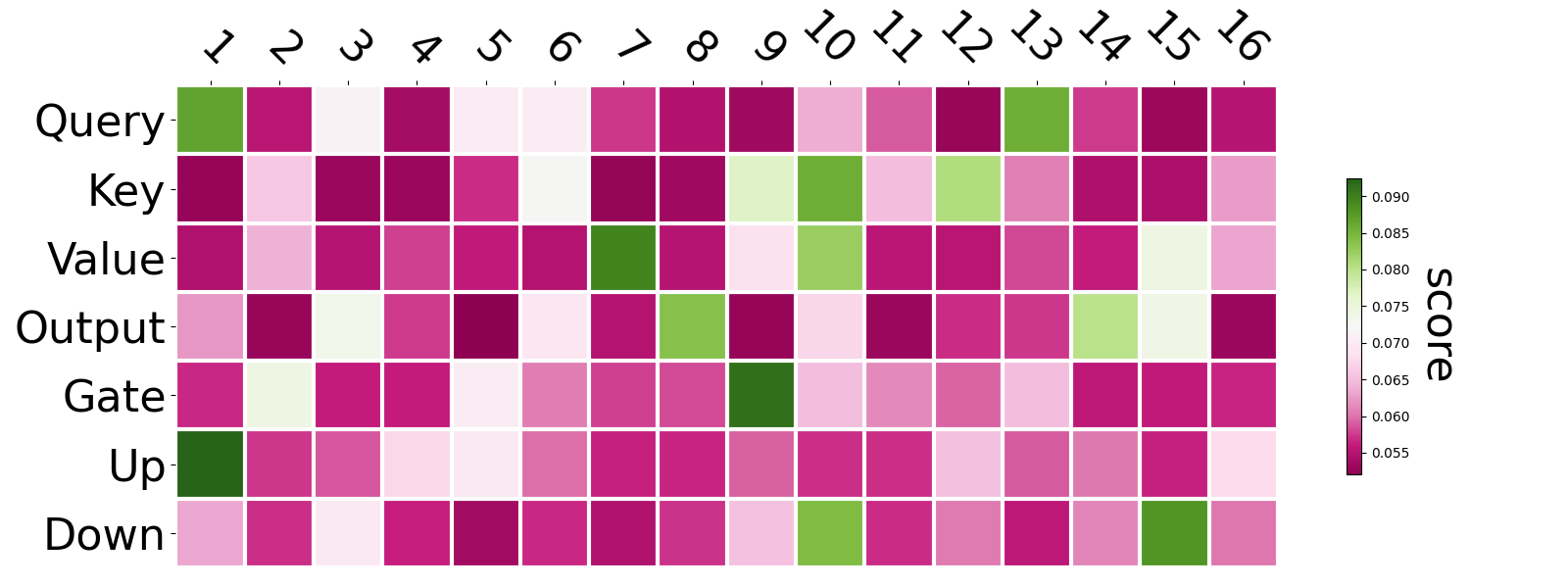}
    \caption{32nd layer}
    \label{subfig:piqa_llama3_8b__layer_31_score_dist}
\end{subfigure}

\caption{Shapley sensitivity importance scores of LoRA ranks on the 8th, 16th, 24th, and 32nd layers of LLaMA-3 8B after finetuning on PIQA.}
\label{fig:piqa_llama3_8b__lora_importance_scores}
\end{figure*}

In Figure \ref{fig:boolq_llama3_8b__lora_importance_scores} and \ref{fig:piqa_llama3_8b__lora_importance_scores}, we present the LoRA importance scores on LlaMA-3 8B on the BoolQ and PIQA tasks.

\begin{figure*}[t]
\centering

\begin{subfigure}{0.96\textwidth}
    \centering
    \includegraphics[width=\linewidth]{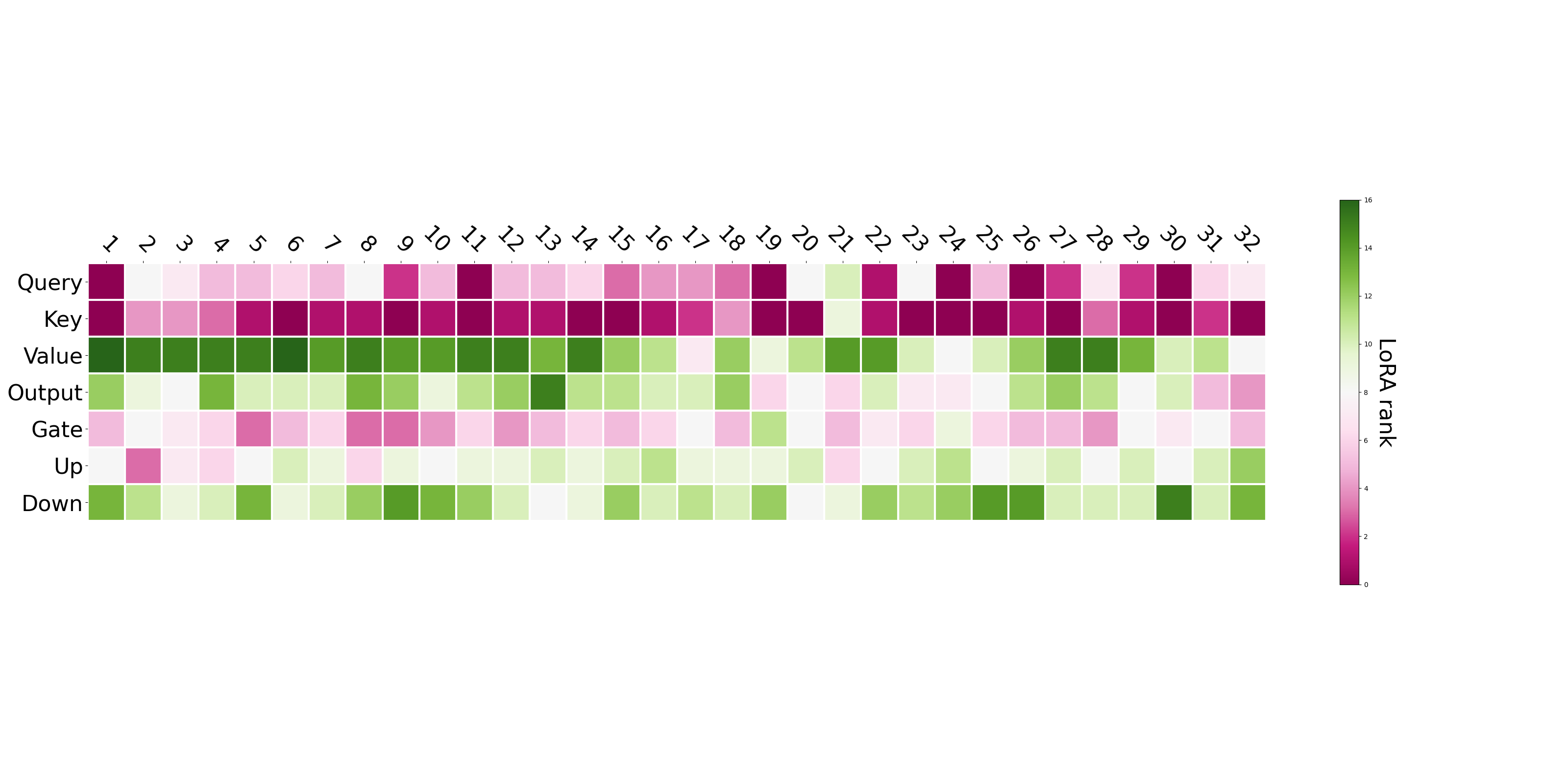}
    \caption{BoolQ}
    \label{subfig:rank_allocation_boolq}
\end{subfigure}

\begin{subfigure}{0.96\textwidth}
    \centering
    \includegraphics[width=\linewidth]{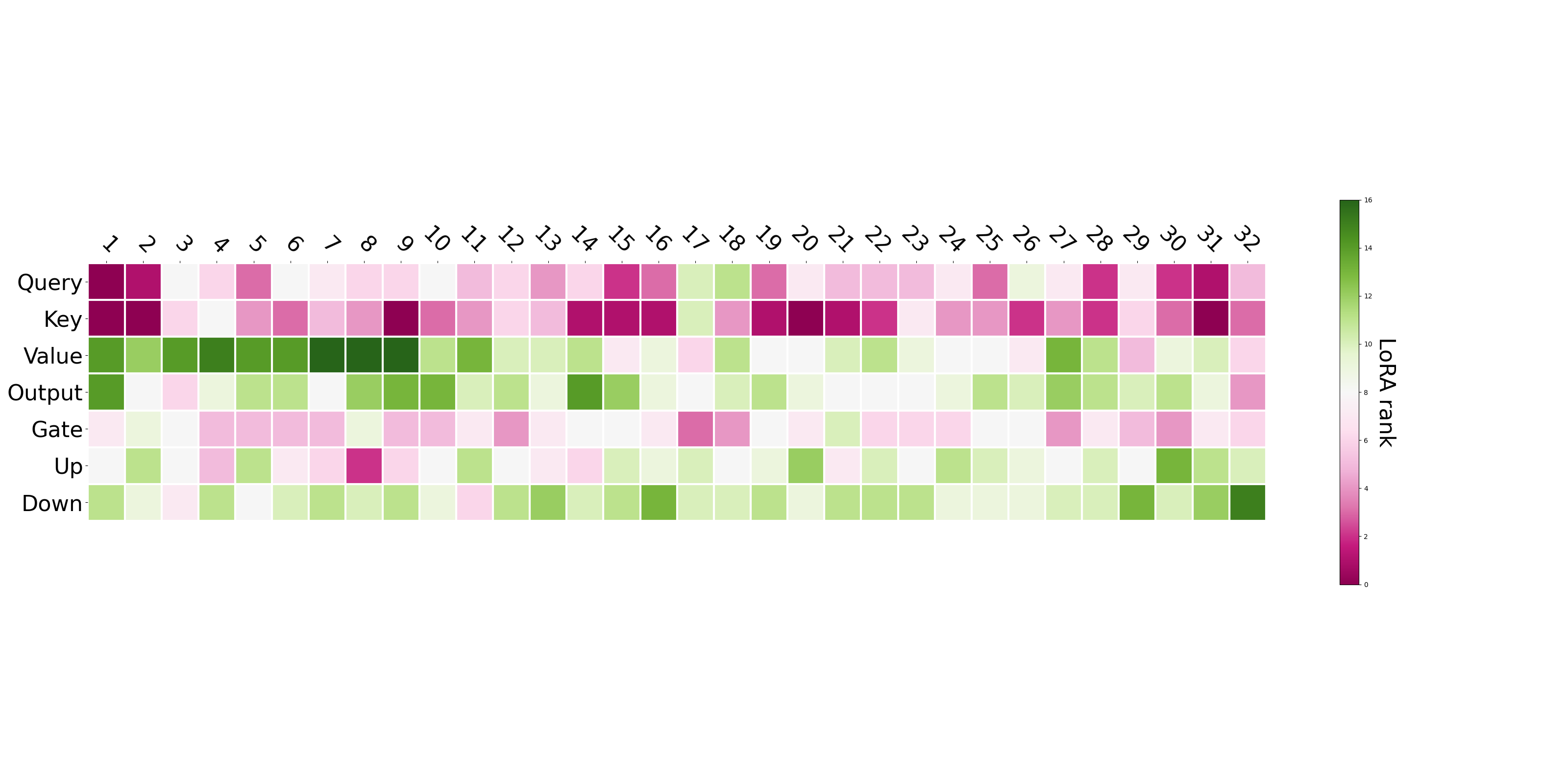}
    \caption{PIQA}
    \label{subfig:rank_allocation_piqa}
\end{subfigure}

\caption{ShapLoRA's LoRA rank allocation results on the LLaMA-3 8B backbone after finetuning on BoolQ and PIQA.}
\label{fig:lora_allocation_results}
\end{figure*}

In Figure \ref{fig:lora_allocation_results}, we present our ShapLoRA's rank allocation results when fine-tuning LlaMA-3 8B on the BoolQ and PIQA tasks.

\end{document}